# FM-AE: Frequency-masked Multimodal Autoencoder for Zinc Electrolysis Plate Contact Abnormality Detection

Canzong Zhou[1], Can Zhou[1], Hongqiu Zhu[1], Wentao Yu[1], Tianhao Liu[1, *]

1. School of automation, Central South University, Changsha 410000
E-mail: zhoucanzong@163.com, zhoucan@csu.edu.cn, hqcsu@csu.edu.cn, yuwentao@csu.edu.cn, tianhaoliu@csu.edu.cn

**Abstract:** Zinc electrolysis is one of the key processes in zinc smelting, and maintaining stable operation of zinc electrolysis is an important factor in ensuring production efficiency and product quality. However, poor contact between the zinc electrolysis cathode and the anode is a common problem that leads to reduced production efficiency and damage to the electrolysis cell. Therefore, online monitoring of the contact status of the plates is crucial for ensuring production quality and efficiency. To address this issue, we propose an end-to-end network, the Frequency-masked Multimodal Autoencoder (FM-AE). This method takes the cell voltage signal and infrared image information as input, and through automatic encoding, fuses the two features together and predicts the poor contact status of the plates through a cascaded detector. Experimental results show that the proposed method maintains high accuracy (86.2%) while having good robustness and generalization ability, effectively detecting poor contact status of the zinc electrolysis cell, providing strong support for production practice.

## 1 Introduction

Zinc electrolysis is one of the key processes in zinc smelting, accounting for more than 90% of the world's zinc production [1]. In the electrolysis process, the energy consumption of zinc electrolysis accounts for more than 75% of the total energy consumption of the zinc smelting process. During zinc electrolysis, the contact quality between the electrode and the busbar may decrease due to improper operation, plate deformation, impurities, corrosion, and other reasons, resulting in contact anomalies. Contact anomalies reduce the contact area at the contact point, increase the contact resistance, and cause local overheating, which decreases the current efficiency of zinc electrolysis. Therefore, rapid detection and treatment of contact anomalies are crucial for the zinc electrolysis process.

Currently, there are two main methods for detecting contact anomalies, based on voltage [2] and temperature [3], respectively. The voltage-based method monitors the state of the contact anomalies by detecting the voltage between adjacent anode-cathode pairs, i.e., the electrolysis cell voltage, which increases with the increase of the contact resistance. The advantage of this method is that the voltage detection sensor is sensitive and the voltage detection accuracy is high. However, it can easily result in false positives due to fluctuations in cell voltage caused by special circumstances such as workers walking on the electrode or cleaning the electrode with a water gun. Moreover, the cell voltage data can be subject to drift and distortion due to voltage sampling and transmission, which affects the accuracy of the detection result over a long monitoring period. The temperature-based method detects contact anomalies by monitoring the temperature at the contact point, which increases with the increase of the contact resistance. The most common method is to use an infrared camera to obtain an infrared image of the entire cell surface, and then use thresholding and other methods to detect anomalies. The advantage of this method is that it has a wide detection range and low cost. However, it is easily affected by external heat sources, such as high-temperature acid mist, which can cause detection errors.

In recent years, researchers have increasingly used deep learning methods for contact anomaly detection in zinc electrolysis, leveraging the superior performance of deep learning methods in complex pattern recognition. For example, the convolutional neural network (CNN)-based method [4] detects and classifies images to determine the contact state of the electrode, while the recurrent neural network (RNN)-based method [5] models and analyzes the cell voltage time series to detect and classify anomalies. These methods have improved the detection accuracy of contact anomalies to some extent. However, they still cannot avoid the fundamental defect of being easily affected by interference from the single-modal data used.

We propose a solution to this problem by using multimodal information for contact anomaly detection in zinc electrolysis. By using image information, we can better reflect the contact area in the cell, while using cell voltage time series information can better reflect the temporal changes in contact anomalies, achieving complementary information. For example, when workers walk on the electrode causing cell voltage disturbance, the temperature information of the electrode is not significantly affected, while when the infrared camera is obstructed by acid mist, the voltage detection sensor is not affected.

To address the problem of modal information complementarity in contact anomaly detection, we propose a frequency-masked multimodal autoencoder (FM-AE), which unifies image information (infrared images) and time series information (cell voltage signals) through auto-encoding, fully excavates the intrinsic features of different modal data, and achieves information complementarity between the two modalities, improving the accuracy of contact anomaly detection in zinc electrolysis and supporting the normal operation of zinc electrolysis.

*This work is supported by the National Key R&D Program of China under Grant No.2022YFE0125000 and in part by the National Natural Science Foundation of China under Grant 92167105 and Natural Science Foundation of Hunan Province of China under Grant 2023JJ10081.

In summary, our work includes:

(1) We proposed a multimodal autoencoder model to unify the representation of image and sequence information.

(2) We trained the proposed multimodal autoencoder model using both image and sequence data and implemented the detection of zinc electrolysis plate contact abnormalities.

(3) We evaluated the performance of our model on a real-world dataset and demonstrated that our model can learn multimodal features and outperform state-of-the-art single-modal models on multiple evaluation metrics.

## 2 Related Works

### 2.1 Voltage-based Methods

Voltage-based methods have been extensively studied for monitoring the working status and energy consumption of electrolytic cells. However, most research has focused on detecting short circuits of the electrode plates, and research on detecting poor contacts is still lacking.

Various methods have been proposed to detect short circuits of the electrode plates. Morales et al. [6] proposed a method for detecting short circuits of electrode plates by placing a voltage sensor between the cathode plate and the conductive busbar and calculating the voltage slope using a sliding window. Cao et al. [2] proposed a method based on the online detection of aluminum cell voltage data, which uses a sliding window and K-means clustering for detecting local anomalies of electrode plates. Furthermore, Man et al. [7] proposed an intelligent diagnosis system for electrolytic cell operation conditions by mining historical data of the aluminum electrolysis monitoring system, which calculates the voltage swing value, voltage swing average, input power, voltage fluctuation, and predicts the alumina concentration value. The system uses expert rules to judge the cell status, but it requires the expertise and knowledge of domain experts to define the rules, which is time-consuming and prone to rule omission in complex scenarios. Some studies have also proposed methods to extract cell status features from the instantaneous voltage curve using wavelet packet decomposition and reconstruction algorithms [8], or by observing the equivalent series resistance (ESR) of the electrolytic cell and combining it with an artificial neural network for cell state diagnosis [9]. However, these methods are susceptible to the influence of plate temperature, which can affect their detection accuracy.

We have noticed that the voltage of the electrode plates can also be influenced by factors such as worker movement and rinsing of the plates, and therefore the use of only voltage signals as detection indicators may not ensure the accuracy of poor contact detection. Our method differs from these methods by integrating the information from both images (infrared images) and sequences (cell voltage signals), which can supplement the cell voltage signal when it is disturbed, thereby improving the accuracy and stability of poor contact detection.

### 2.2 Temperature-based Methods

Temperature-based methods have also been explored for detecting the working state of electrodes in an electrolytic cell. However, research on using temperature as an indicator for detecting contact anomalies is still lacking. There are two main types of temperature-measuring instruments: temperature sensors and infrared cameras.

Methods based on temperature sensors require a large number of sensors to be placed on the surface of the cell. Han X et al. [11] used an RS-485 network to transmit temperature sensor data for the detection of short circuits in the electrolytic cell electrodes. However, the large size of the sensors makes them susceptible to damage during operation, making long-term stable operation difficult. Aqueveque P et al. [12] proposed an in-cell temperature sensor that transmits temperature data to a host computer via Bluetooth. This sensor is small in size and operates stably, making it possible to quickly track changes in electrode temperature. However, due to cost and communication bandwidth limitations, it is difficult to detect a large number of electrolytic cell electrodes simultaneously.

Methods based on infrared cameras acquire an infrared image that represents the entire surface of the electrolytic cell. As a non-contact temperature measurement method, it has the advantages of a wide temperature measurement area and high real-time performance. Jia R et al. [13] proposed a pixel-sorting PCA feature extraction algorithm to obtain sample features and then used a support vector machine classifier to identify short circuits in the electrodes. Li X et al. improved the Difference of Gaussian (DoG) filter to match the grayscale distribution of infrared images and achieved the detection of electrode short circuits by estimating the periodicity of the texture in the second stage. In recent years, some researchers have begun to use deep learning methods to evaluate electrode contact anomalies. Zhu H et al. [3,14] have used Mask R-CNN-based object recognition and LSTM-based sequence prediction methods to evaluate electrode status.

However, it is difficult to detect anomalous plates quickly, accurately, and completely using infrared images due to irregular coverage, acid mist, water vapor diffusion, and other complex factors. Our method differs from these methods in that we also use the voltage signal from the electrolytic cell as a supplement when the infrared image is disturbed by acid mist and other factors, improving the accuracy and stability of detecting electrode contact anomalies.

### 2.3 Multimode Methods

Multimodal research refers to a research method that combines data from different modalities, such as images, speech, text, etc., to obtain more comprehensive and accurate information. In many fields, including computer vision [15], natural language processing [16], and human-computer interaction [17], multimodal research has become a hot research direction. This is because multimodal research can provide information from different perspectives, thereby obtaining more comprehensive and accurate results. That is, in a multimodal representation that contains similar information, a multimodal representation can still be generated to obtain more information when some modal data information is missing [18].

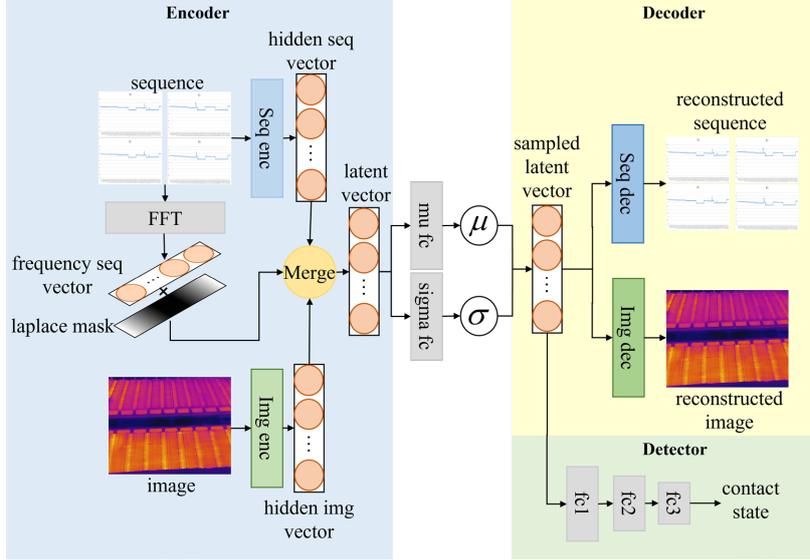

Fig. 1: FM-AE, which consists of three parts: encoder, decoder, and detector.

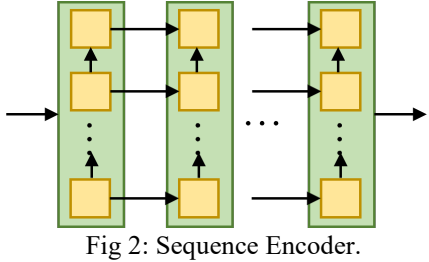

Fig 2: Sequence Encoder.

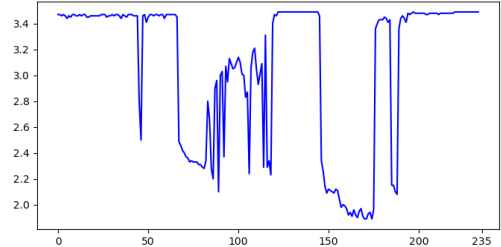

Fig 3: Voltage sequence in poor contact condition.

Kiros et al. [19] built a neural network that can learn modal co-acting semantic representations and is expected to evaluate the performance of the co-acting semantic representation by observing the reconstruction ability of the co-acting semantic representation to the original input data of each modality. Mroueh et al. [20] cascaded the representations of sound and visual inputs learned by neural networks and generated prediction results based on the co-acting semantic representation generated by the cascade. Kim et al. [21] used a similar deep Boltzmann machine to fuse visual and auditory modalities in audiovisual emotion recognition to generate a joint representation. Hori et al. [15] extracted image and audio features from videos respectively and used multimodal attention to build an end-to-end video dialogue model, proving that multimodal features can improve the quality of generation tasks in dynamic scenes (videos). Li et al. [16] proposed a general encoder structure that can learn joint representations of vision and language in a pre-training manner. Khattar et al. [22] proposed an end-to-end multimodal variational autoencoder (MVAE), which fused features of text and images and coupled with a binary classifier to perform fake news detection tasks.

We consider that the multimodal model can combine different information channels to understand tasks well. We try to build an autoencoder that integrates and fuses the features of both image and sequence modalities, find a common semantic space for unified representation, i.e., the encoded feature vector, and use this feature vector for predicting contact anomalies.

## 3 Network

### 3.1 FM-AE

Our proposed FM-AE (Frequency-masked Multimodal Autoencoder), as shown in Fig. 1, is based on the Variational Autoencoder architecture [22], and consists of three main components: an encoder, a decoder, and a detector. The encoder encodes the voltage sequence and the infrared image into a latent vector; the decoder decodes the latent vector to reconstruct the original voltage sequence and infrared image; and the detector performs contact anomaly detection based on the encoding results, i.e., the latent vector.

### 3.2 Encoder

The encoder is mainly responsible for learning a unified representation of the sequence and image modalities. It consists of two parts, the sequence encoder, and the image encoder. Its inputs are a $60 \times 16$ voltage sequence and a $640 \times 512 \times 3$ infrared image. In the voltage sequence input, the rows represent the sequence length, with a total of 60 sampling points, a sampling period of 10 seconds, and a total cycle duration of 10 minutes. The columns represent the number of electrolytic cells, with 16 electrolytic cells in total, corresponding to one column. The same 16 electrolytic cells are also included in each infrared image.

To extract temporal features from the voltage sequence, we use 64 consecutive stacked unidirectional Long Short Term Memory networks (LSTMs) in the sequence encoder, as shown in Fig. 2. Because LSTMs introduce gate mechanisms (such as input gates, forget gates, and output gates) to control information flow and processing, the state at each moment includes not only the current input but also the previous state and previous output, making it superior to traditional RNNs in processing long sequence data [23]. The input sequence is gradually processed, and the sequence's autocorrelation features are obtained through gate mechanisms in multiple iterative updates at different time steps, and the encoded $1 \times 64$ hidden sequence vector is finally output.

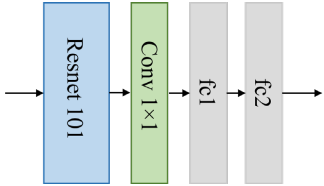

Fig 4: Image encoder

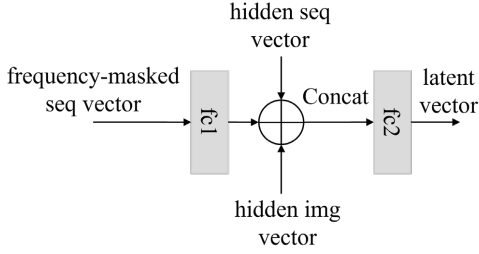

Fig 5: Merge module

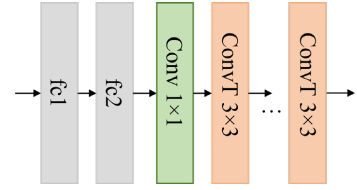

Fig 6: Image decoder

At the same time, we designed a frequency-masked mechanism to help the encoder extract features better. We found that when there is poor contact, there is high-frequency jitter in the voltage frequency, as shown in Fig. 3. Therefore, we first perform a fast Fourier transform (FFT) on the voltage sequence to obtain its frequency domain representation. Then, we move the first column (DC signal) of the spectrum to the middle position, and use a Laplace mask (0 in the middle, gradually increasing to 1 on both sides) to multiply and superimpose the frequency domain representation, retaining the high-frequency signal as additional feature information.

In various visual or image tasks, Convolutional Neural Networks (CNNs) have performed well. Therefore, we adopted the CNN architecture as the image encoder, as shown in Fig. 4. We also noticed the excellent performance of ResNet in image feature extraction tasks, so we used pre-trained Resnet 101 to extract image features, which not only reduces training costs but also achieves good results. After Resnet101, we use a $1 \times 1$ convolution to implement information flow between channels, and then reshape the feature into a $1 \times 16$ hidden sequence vector through two fully connected layers (FC).

At the end of the encoder, as shown in Fig. 5, we use a Merge module and a FC to reshape the frequency-masked seq vector into a $1 \times 64$ vector, and then concatenate the hidden seq vector, frequency-masked seq vector, and hidden image vector. Finally, we output a $1 \times 64$ latent vector through another FC layer.

### 3.3 Decoder

We follow the approach of MVAE [22] and use two FC layers to obtain $\mu$ and $\sigma$, which can be seen as the mean and variance of the distribution under the multimodal unified representation. Next, we re-sample a variable under this multimodal distribution using re-parameterization. The multimodal unified representation $R_m$ after re-parameterization is:

$$R_m = \mu + \sigma \bullet \varepsilon \qquad (1)$$

where $\varepsilon$ is a randomly sampled variable from this distribution.

The structure of the decoder is similar to that of the encoder, with the difference being that the decoder is responsible for upsampling and reconstructing the data. Similarly, the decoder is also mainly composed of two parts: the sequence decoder and the image decoder. The structure of the sequence decoder is almost identical to that of the sequence encoder, with 64 consecutive stacked unidirectional LSTMs used to decode $R_m$, and a final FC used to reshape the features into a $60 \times 16$ voltage sequence.

In the image decoder, as shown in Fig. 6 we first use two FC layers to expand $R_m$, followed by a 1x1 convolution to fuse information across channels. Then, we use five $3 \times 3$ transposed convolutions to perform upsampling and reconstruct the image. We use a cascade of three FC layers as the detection module in FM-AE, which outputs a binary classification result of $1 \times 1$ indicating the presence or absence of electrode contact anomalies, where 1 indicates the presence of an anomaly and 0 indicates normally.

### 3.4 Loss function

The loss function of the model can be represented as:

$$\mathcal{L} = \alpha \bullet \mathcal{L}_{recseq} + \beta \bullet \mathcal{L}_{recimg} + \gamma \bullet \mathcal{L}_{kl} \qquad (2)$$

$$\mathcal{L}_{recseq} = \mathbb{E}_{m \sim M} \left[ \sum_{i=1}^{n_s} \sum_{c=1}^{C} \left( \hat{r}_s^{(i)} - r_s^{(i)} \right)^2 \right] \qquad (3)$$

$$\mathcal{L}_{recimg} = \mathbb{E}_{m \sim M} \left[ \sum_{i=1}^{n_p} \left( \hat{r}_p^{(i)} - r_p^{(i)} \right)^2 \right] \qquad (4)$$

$$\mathcal{L}_{kl} = \frac{1}{2} \sum_{i=1}^{n_m} \left( \mu_i^2 + \sigma_i^2 - \log(\sigma_i) - 1 \right) \qquad (5)$$

where $\mathcal{L}$ is the total loss, $\mathcal{L}_{recseq}$ is the reconstruction loss for the sequence, $\mathcal{L}_{recimg}$ is the reconstruction loss for the image, and $\mathcal{L}_{kl}$ is the KL divergence between the latent vector and a normal distribution, weighted by $\alpha$, $\beta$, and $\gamma$ respectively. $M$ is the multimodal dataset, $n_s$ is the number of samples for the sequence (i.e., number of cells), $C$ is the sequence length, $n_p$ is the number of pixels in the image, and $n_m$ is the dimensionality of the latent vector.

Table 1: Comparison of detection performance between FM-AE and other methods

|           | LSTM   | GRU    | Resnet 101 | FM-AE (Ours) |
|-----------|--------|--------|------------|--------------|
| Accuracy  | 0.7622 | 0.7434 | 0.6763     | 0.8623       |
| Precision | 0.5231 | 0.5537 | 0.6003     | 0.6373       |
| Recall    | 0.5526 | 0.5503 | 0.5025     | 0.9769       |
| F1-score  | 0.5374 | 0.5520 | 0.5471     | 0.7714       |
| AUC       | 0.7304 | 0.6871 | 0.6504     | 0.9108       |

## 4 Experiments

We trained and tested our proposed FM-AE model using actual data collected from a zinc smelting plant, which includes matched voltage sequences and infrared images with corresponding labels for plate contact anomalies. Firstly, we performed unsupervised training on the matched voltage sequences and infrared images using the FM-AE model. After convergence, we froze the network weights of FM-AE and cascaded the detector on top, and continued training the detector using the labeled data. As shown in Table 1, FM-AE outperforms the state-of-the-art single-modal detection methods on all metrics, which demonstrates that FM-AE can effectively integrate multiple modalities and achieve better performance in complex anomaly detection tasks.

## 5 Conclusion

We proposed a multimodal autoencoder-based method for detecting zinc electrolysis cell plate contact anomalies. This method unified the representations of infrared images and electrolysis cell voltage signals in a shared semantic space. Experimental results showed that the proposed method had high accuracy and stability in detecting plate contact anomalies in zinc electrolysis cells. In addition, we compared the performance of the model using different input modalities and found that the fusion of multimodal information could improve the model's performance. In summary, this study provides a new approach and method for detecting plate contact anomalies in electrolysis cells, which has practical value and significance for industrial production.